\documentclass[11pt]{article}

\usepackage[preprint]{acl}

\usepackage{times}
\usepackage{latexsym}
\usepackage[T1]{fontenc}
\usepackage[utf8]{inputenc}
\usepackage{microtype}
\usepackage{inconsolata}
\usepackage{graphicx}

\usepackage{amsmath,amssymb}
\usepackage{makecell}
\usepackage{multirow}
\usepackage{booktabs}
\usepackage{bm}
\usepackage{amsfonts}
\usepackage{booktabs}
\usepackage{subcaption}
\usepackage{graphicx}

\usepackage{pifont}
\newcommand{\yesmark}{\ding{51}}  % Check mark
\newcommand{\nomark}{\ding{55}}  % Cross mark
\newcommand{\ms}[2]{{#1\tiny{$\pm$#2}}}
\usepackage{xcolor}

\newcommand{\bd}[1]{{#1}}

\usepackage{colortbl}
\usepackage[table]{xcolor}
\usepackage{verbatim}
\usepackage[normalem]{ulem}
\useunder{\uline}{\ul}{}
\usepackage{tcolorbox}

\usepackage[normalem]{ulem}

\title{Autonomous Chain-of-Thought Distillation for Graph-Based Fraud Detection}

\author{
  Yuan Li\textsuperscript{1}\thanks{Work done during internship at ByteDance.} \quad
  Jun Hu\textsuperscript{1} \quad
  Bryan Hooi\textsuperscript{1} \quad
  Bingsheng He\textsuperscript{1} \quad
  Cheng Chen\textsuperscript{2} \\
  \textsuperscript{1}National University of Singapore\\
  \textsuperscript{2}ByteDance Inc.\\
  \texttt{li.yuan@u.nus.edu} \quad
  \texttt{\{jun.hu, dcsbhk, dcsheb\}@nus.edu.sg} \\
  \texttt{chencheng.sg@bytedance.com}
}

\begin{document}
\maketitle
\begin{abstract}
Graph-based fraud detection on text-attributed graphs (TAGs) requires jointly modeling rich textual semantics and relational dependencies. However, existing LLM-enhanced GNN approaches are constrained by predefined prompting and decoupled training pipelines, limiting reasoning autonomy and weakening semantic–structural alignment. We propose FraudCoT, a unified framework that advances TAG-based fraud detection through autonomous, graph-aware chain-of-thought (CoT) reasoning and scalable LLM–GNN co-training. To address the limitations of predefined prompts, we introduce a fraud-aware selective CoT distillation mechanism that generates diverse reasoning paths and enhances semantic–structural understanding. These distilled CoTs are integrated into node texts, providing GNNs with enriched, multi-hop semantic and structural cues for fraud detection. Furthermore, we develop an efficient asymmetric co-training strategy that enables end-to-end optimization while significantly reducing the computational cost of naive joint training. Extensive experiments on public and industrial benchmarks demonstrate that FraudCoT achieves up to 8.8\% AUPRC improvement over state-of-the-art methods and delivers up to 1,066$\times$ speedup in training throughput, substantially advancing both detection performance and efficiency.
\end{abstract}

\section{Introduction}

\textbf{Graph-based fraud detection} on e-commerce and financial platforms typically uses a graph to model entities such as users, items, and transactions (as nodes) and relations between them (as edges), where both node attributes and graph structure are essential for identifying fraudulent patterns~\citep{dou2020enhancing, liu2021pcgnn}.
In many scenarios, node attributes are rich textual content—such as product reviews—forming text-attributed graphs (TAGs).
\textbf{Fraud detection on TAGs} has emerged as an important research area, as it requires simultaneously addressing complex textual encoding and exploiting relational information~\citep{yang2025flag, huang2025can}.
Fraud detection on TAGs is particularly challenging because fraudulent signals in text are often subtle and implicit, requiring deeper semantic reasoning rather than surface-level pattern matching.
Various Graph Neural Networks (GNNs) have been proposed for this task~\citep{dou2020enhancing,zhang2021fraudre,tang2022rethinking}.

\begin{figure}[!t]
\centering
\begin{subfigure}{\linewidth}
    \centering
    \includegraphics[width=\linewidth]{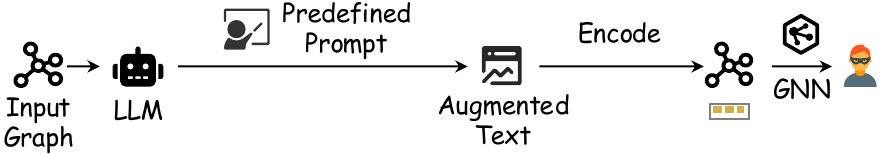}
    \caption{Existing LLM-enhanced GNNs rely on predefined prompts for text augmentation, limiting autonomous reasoning.}
    \label{fig:analysis-predefined}
    \vspace{1em}
\end{subfigure}
\begin{subfigure}{\linewidth}
    \centering
    \includegraphics[width=\linewidth]{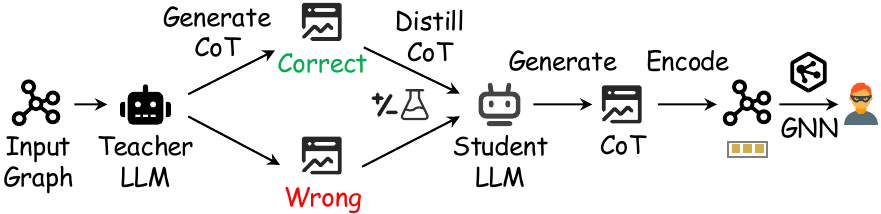}
    \caption{Our proposed method distills chain-of-thought reasoning from teacher LLM to student for free-form graph reasoning.}
    \label{fig:comp-text}
\end{subfigure}
\caption{Comparison between predefined-prompt augmentation and our graph reasoning approach.}
\vspace{-3mm}
\label{fig:analysis-cot}
\end{figure}

\textbf{LLM-enhanced GNNs}. 
Early approaches to fraud detection on TAGs typically used Language Models (LMs) such as BERT~\citep{devlin2019bert} to encode node text into fixed embeddings, which were then processed by GNNs. 
To improve textual representations, recent work has begun leveraging Large Language Models (LLMs) to enrich raw text before feeding it to LMs, giving rise to LLM-enhanced GNNs~\citep{he2024harnessing, yang2025flag, huang2025can, liu2025scalable}.
Figure~\ref{fig:analysis-predefined} shows a typical LLM-enhanced GNN approach: LLMs process raw node text to generate enriched textual representations, which are then encoded by LMs and processed by GNNs alongside graph structure.
The enrichment can take various forms, such as extracting task-relevant features~\citep{he2024harnessing} or generating high-level semantic summaries~\citep{huang2025can}.

\begin{figure}[!t]
  \centering
  \begin{subfigure}[b]{0.4\linewidth}
    \centering
    \includegraphics[width=\linewidth]{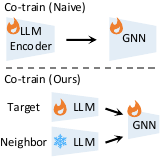}
    \caption{Co-train paradigms.}
    \label{fig:cotrain-methods}
  \end{subfigure}
  \hfill
  \begin{subfigure}[b]{0.57\linewidth}
    \centering
    \includegraphics[width=\linewidth]{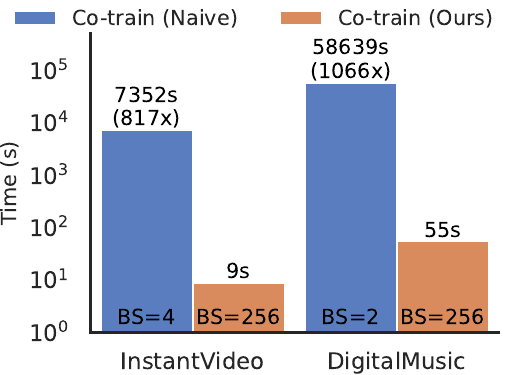}
    \caption{Training time per epoch.}
    \label{fig:cotrain-speedup}
  \end{subfigure}
  \caption{Training efficiency between naive co-training and our proposed asymmetric co-training. By performing asymmetric target and neighbor encoding, FraudCoT achieves reasonable training cost while maintaining joint optimization.}
  \vspace{-3mm}
  \label{fig:comparison-cotrain}
\end{figure}

Despite these promising advances, existing LLM-enhanced GNNs face two major limitations. 
(1) \textbf{Lack of autonomous reasoning due to predefined prompting.}
As shown in Figure~\ref{fig:analysis-predefined}, most current methods rely on manually designed extraction rules or predefined templates to guide LLMs in generating auxiliary information~\citep{he2024harnessing, yang2025flag}
—for example, instructing the LLM to "extract sentiment and keyword features" or "identify specific fraud indicators."
Such prompting strategies restrict LLMs to shallow pattern matching rather than multi-step reasoning, preventing them from uncovering subtle multi-hop dependencies—like coordinated accounts connected through shared devices or intermediary transactions—that often underlie fraudulent behaviors.
(2) \textbf{Weakened semantic--structural alignment due to decoupled training.}
End-to-end training enables semantic-structural alignment, where node representations learned by the LLM are optimized to work effectively with structure-based operations (e.g., neighbor aggregation) in GNNs. 
However, prior work~\citep{liu2025scalable, yang2025flag, he2024harnessing} typically degrades end-to-end optimization into a decoupled paradigm where LLM-based semantic encoding and GNN-based structural learning are trained separately. Such separation is problematic in fraud detection, where signals often arise from joint text–graph modeling, e.g., when benign-looking texts become suspicious once linked through a lot of shared devices.

We address these limitations by introducing FraudCoT, a unified framework that advances LLM-enhanced GNNs through graph-aware CoT reasoning and efficient semantic--structural co-training.  
(1) To enable autonomous reasoning beyond predefined prompting~\citep{he2024harnessing,yang2025flag}, we design a \emph{fraud-aware selective CoT distillation} mechanism that allows the teacher LLM to freely explore both semantic and structural cues and generate diverse reasoning paths. These paths include both correct and incorrect trajectories, enabling the student LLM to learn robust graph reasoning through positive and negative distillation.  
(2) We integrate the distilled CoTs into node texts to form CoT-augmented representations, providing downstream GNNs with enriched multi-hop semantic and structural cues.  
(3) To enable end-to-end LLM--GNN optimization for better semantic-structural alignment, we develop an \emph{efficient asymmetric co-training} strategy that significantly reduces the computational cost of naive joint training.
By decoupling LLM encoding for target and neighbor nodes, this design reduces the complexity of LLM inference from exponential in the number of hops to constant, achieving up to \textbf{1,066$\times$} speedup.

Our contributions are summarized as follows:
\begin{itemize}
    \item We propose FraudCoT, a CoT-enhanced framework for graph-based fraud detection. To enable autonomous reasoning beyond predefined prompting, we introduce a fraud-aware selective CoT distillation mechanism that produces diverse reasoning paths for more reliable graph reasoning.
    \item To enable end-to-end LLM--GNN optimization for better semantic-structural alignment, we develop an efficient asymmetric co-training strategy that significantly reduces the computational cost of naive joint training.
    \item Extensive experiments on public and industrial fraud detection benchmarks demonstrate that FraudCoT achieves up to 8.8\% improvement in AUPRC over state-of-the-art baselines, while delivering orders-of-magnitude speedup in training throughput.
\end{itemize}

\section{Related Work}

\subsection{Graph-based Fraud Detection}

Graph-based fraud detection methods exploit relational structures to capture collective suspicious patterns that are invisible from isolated instances. 

Early work designs GNN architectures that are robust under class imbalance and noisy neighborhoods: PC-GNN~\citep{liu2021pcgnn} introduces label-aware subgraph sampling and neighbor re-weighting to improve minority-class recall on industrial fraud graphs, while H2-FDetector~\citep{shi2022h2fdetector} explicitly models both homophilic and heterophilic relations via relation-aware aggregation so that informative neighbors can be leveraged even when fraudsters connect to benign accounts. 

Subsequent methods focus on stability and scalability in large, noisy graphs. 
ConsisGAD~\citep{chen2024consistency} encourages prediction consistency under neighborhood perturbations to mitigate the impact of spurious edges; 
PMP~\citep{zhuo2024partitioning} proposes partitioned message passing that distinguishes neighbors by class and applies adaptive aggregation to balance homophilic and heterophilic signals;
GAAP~\citep{duan2025global} combines global and local attention mechanisms to better aggregate heterogeneous relational signals. 
Although these GNN-based approaches effectively encode structural information, they typically rely on shallow or fixed textual features, which limits their ability to exploit rich semantics in text-heavy fraud detection scenarios such as e-commerce reviews and promotion-abuse campaigns.

\subsection{LLM-enhanced GNNs for Fraud Detection}

To fully utilize textual attributes on graphs, a recent line of work integrates large language models with GNNs and gives rise to LLM-enhanced GNNs for fraud detection. TAPE~\citep{he2024harnessing} prompts an LLM to summarize reviews and generate task-specific analyses, which are then encoded as node features for a downstream GNN; STAR~\citep{liu2025scalable} and FLAG~\citep{yang2025flag} similarly use LLMs to extract discriminative textual cues or pseudo-labels and train GNNs on the resulting semantic representations. These methods demonstrate that LLM-based semantics significantly boost detection performance over purely structural GNNs, but they usually rely on predefined prompts and adopt decoupled training, which restricts autonomous reasoning and weakens semantic–structural alignment. 

As summarized in Table~\ref{tab:llm-gnns}, existing LLM-enhanced GNNs either rely on predefined rule-based prompts or decoupled training, which limits reasoning autonomy and semantic-structural alignment, respectively. Our work bridges these gaps through graph-aware CoT distillation and efficient co-training that decouples target and neighbor updates. This yields autonomous reasoning aligned with graph-level objectives under realistic budgets.

Beyond this line, there is a broader family of graph-enhanced LLMs that inject structural bias into language models, such as GraphGPT and HiGPT~\citep{tang2024graphgpt,tang2024higpt}, as well as graph-aware instruction-tuned models like LLaGA and InstructGLM~\citep{ye2023language}. Other works attach lightweight GNN adapters to frozen LLMs~\citep{huang2024graphadapter} or alternate between updating the LLM and GNN in an EM-style scheme~\citep{zhao2023glem}. Our work falls into the LLM-enhanced GNN category and is complementary to graph-enhanced LLMs: instead of using fixed prompts or fully generative graph reasoning, we distill free-form chain-of-thought reasoning from a teacher LLM into a student and couple it with an efficient co-training scheme, enabling autonomous graph-aware reasoning and end-to-end optimized LLM–GNN representations for fraud detection.

\subsection{Graph Reasoning and Distillation}

Chain-of-Thought (CoT) prompting and its training-time variants have proven effective for complex reasoning. Self-consistency has been proposed to sample diverse rationales and aggregate them by voting \citep{wang2023selfconsistency}, while rationale distillation from teacher LLMs has been shown to endow smaller students with strong step-by-step reasoning \citep{magister2023teaching,hsieh2023distill}. However, applying such reasoning to graphs introduces two gaps: (i) most LLM+GNN works for TAGs use templated prompts or explanation-as-features \citep{he2024harnessing}, which constrain autonomy for multi-hop, cross-neighborhood inference; and (ii) frequent on-the-fly LLM reasoning is computationally expensive on sampled subgraphs. Our proposed method addresses these challenges by distilling free-form CoT reasoning from a large-scale teacher LLM into a small student LLM.

\begin{table}[t]
\centering
\resizebox{\linewidth}{!}{
\begin{tabular}{@{}cccc@{}}
\toprule
Method & Analysis Generation & \makecell{LLM--GNN\\Training} & Efficiency \\ \midrule
Naive Co-train~\citep{liu2025scalable} & None & End-to-end & Low \\
STAR~\citep{liu2025scalable}      & None & Decoupled & High \\
TAPE~\citep{he2024harnessing}      & Predefined analysis & Decoupled & High \\
FLAG~\citep{yang2025flag} & Predefined analysis & Decoupled & High \\
FraudCoT (Ours)  & Free-form reasoning & End-to-end & High \\ 
\bottomrule
\end{tabular}
}
\caption{Comparison of analytical text generation, LLM-GNN training paradigms, and efficiency between FraudCoT and existing LLM-enhanced GNNs.}
\label{tab:llm-gnns}
\end{table}

\section{Methodology}
We propose FraudCoT, a LLM-enhanced GNN framework that unifies chain-of-thought reasoning and LLM-GNN co-training to detect fraudulent entities on text-rich graphs. 

\subsection{Problem Setting}

Let a heterogeneous text-attributed graph be defined as $G = (V, E, \mathcal{R}, \mathcal{X})$, where \(V\) is the set of nodes, \(E \subseteq V \times V \times \mathcal{R}\) the set of typed edges, \(\mathcal{R} = \{r_1, r_2, \dots, r_{|\mathcal{R}|}\}\) denotes relation types, and \(\mathcal{X} = \{x_v\}_{v \in V}\) represents textual attributes (e.g., user profiles, transaction descriptions). Each node \(v\) corresponds to an entity labeled \(y_v \in \{0, 1\}\), indicating whether it is fraudulent. The task of graph-based fraud detection aims to learn a classifier \(f_\theta: V \rightarrow [0, 1]\) to predict the fraudulent scores of each node.

\subsection{Overview of FraudCoT}

\begin{figure*}
    \centering
    \includegraphics[width=1\linewidth]{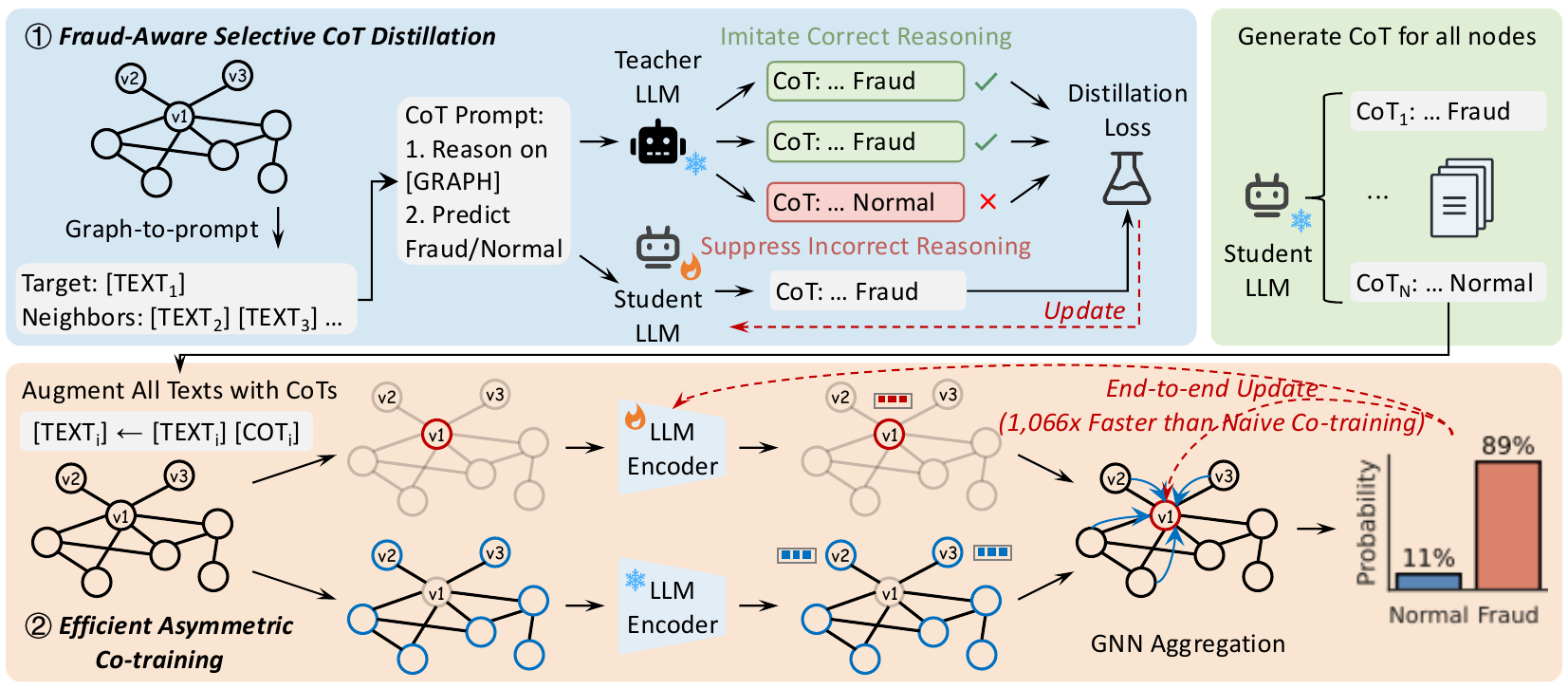}
    \caption{Overview of the proposed FraudCoT.}
    \label{fig:framework}
\end{figure*}

As illustrated in Figure~\ref{fig:framework}, FraudCoT operates in two complementary stages:

\begin{itemize}
    \item \textbf{Fraud-Aware Selective CoT Distillation} involves a teacher LLM performing graph reasoning, generating step-by-step CoTs. A student model is then finetuned to imitate correct reasoning while suppressing erroneous ones, producing concise and faithful reasoning paths for all nodes.
    \item \textbf{Efficient Asymmetric Co-training} encodes the CoT-augmented texts with an LLM encoder and jointly trains it with a GNN classifier. The joint optimization aligns semantic reasoning with structural learning while maintaining computational efficiency.
\end{itemize}

This two-stage design separates reasoning acquisition from representation learning, yet allows both to reinforce each other:  
the distilled CoTs provide graph-grounded semantics, while joint training ensures those semantics remain consistent with relational patterns.

\subsection{Fraud-Aware Selective CoT Distillation}

Existing LLM-enhanced approaches suffer from a lack of autonomous reasoning due to predefined prompting, restricting LLMs to shallow pattern matching rather than genuine multi-step inference. To overcome this limitation, we propose a \emph{fraud-aware selective CoT distillation} module that empowers the model to autonomously extract complex fraud patterns through graph reasoning.

\textbf{Teacher Reasoning.}
For a labeled node \(v_i\) with neighborhood \(\mathcal{N}(v_i)\), we construct a prompt following the template described in Figure~\ref{fig:fraudcot-template}.  
A large teacher LLM, such as DeepSeek-R1~\cite{guo2025deepseek}, generates \(S\) reasoning paths for each node:
\begin{equation}
r_{i}^{T, s} = \text{Teacher}(x_i, \mathcal{N}(v_i)), \quad s = 1, \dots, S
\end{equation}
If its decision matches the true label, each CoT is labeled as positive; otherwise, as negative.  
This yields a set of triples
$
\{(p_i, r_{i}^{T, s}, y_{i}^{T, s})\}_{s=1}^{S}
$
where \(p_i\) is the prompt, \(r_{i}^{T, s}\) the \(s\)-th teacher reasoning, and \(y_{i}^{T, s}\) its correctness label. Note that we only generate teacher CoTs on a sampled set of $U$ nodes, considering the prohibitive cost of running teacher LLMs on the entire graph.

\textbf{Distillation of Reasoning Paths.}
Unlike standard distillation that only transfers output distributions, FraudCoT explicitly distinguishes positive and negative reasoning paths. 
The teacher’s correct CoTs serve as positive trajectories that demonstrate faithful reasoning logic, while incorrect CoTs act as negative trajectories exposing spurious correlations. By learning to imitate the former and suppress the latter, the student acquires a more robust reasoning mechanism—generalizing beyond the teacher’s text style while avoiding its reasoning biases. 
Specifically, given that each node yields \(S\) teacher reasoning paths 
\(\{r_{i}^{T,s}\}_{s=1}^S\), we finetune the student LLM with positive–negative distillation objective:
\begin{equation}
\begin{aligned}
\mathcal{L}_{\text{distill}} = 
\mathbb{E}_{i,s}\Big[ &
    y_{i}^{T,s} \, \mathcal{L}_{\text{CE}}(p_i, r_{i}^{T,s}) \\
&+ \lambda \, (1-y_{i}^{T,s}) \, \mathcal{L}_{\text{UL}}(p_i, r_{i}^{T,s})
\Big]
\end{aligned}
\end{equation}
with
\begin{equation}
\begin{aligned}
\mathcal{L}_{\text{CE}}(p_i, r_{i}^{T,s}) &= 
    -\log P_\theta(r_{i}^{T,s}\mid p_i) \\
\mathcal{L}_{\text{UL}}(p_i, r_{i}^{T,s}) &= 
    -\log\!\bigl(1 - P_\theta(r_{i}^{T,s}\mid p_i)\bigr)
\end{aligned}
\end{equation}
where the cross-entropy loss \(\mathcal{L}_{\text{CE}}\) encourages reproducing correct reasoning, 
while the unlikelihood loss \(\mathcal{L}_{\text{UL}}\) discourages reproducing incorrect reasoning.  
Low-rank adaptation (LoRA)~\citep{hu2022lora} adapters are used for efficient parameter updates.

\textbf{Node-level CoT Generation.}  
After distillation, the student model produces a reasoning path for each node. These enriched texts embed explicit reasoning into node semantics, providing transparent rationales that can be further integrated into structural learning:
\begin{equation}
r_i = \text{Student}(x_i, \mathcal{N}(v_i))
\end{equation}
Each node’s final text concatenates its raw text and CoT reasoning path:
\begin{equation}
\tilde{x}_i = x_i \oplus r_i
\end{equation}

\subsection{Efficient Asymmetric Co-training}

Prior works also exhibit weakened semantic--structural alignment due to decoupled training, where semantic encoding and structural learning are optimized in isolation, failing to capture their mutual dependencies. To address this, we introduce an \emph{efficient asymmetric co-training} strategy that enables end-to-end joint optimization while maintaining computational feasibility.

\textbf{Asymmetric semantic encoding.}
For each training batch $\mathcal{B}$, the LLM encoder processes only the CoT-augmented texts
of each target node $v_i \in \mathcal{B}$:
\begin{equation}
h^{(0)}_i = \mathrm{Enc}_\theta(\tilde{x}_i)
\label{eq:target-encode}
\end{equation}
In contrast, all neighbor nodes $v_j\in\mathcal{N}(v_i)$ use cached embeddings:
\begin{equation}
\bar{h}^{(0)}_j = \mathrm{Cache}(j)
\label{eq:cache}
\end{equation}
which is computed using the initial parameters $\mathrm{Enc}_{\theta_0}$.  
This design preserves CoT-enhanced semantics for all nodes while reducing LLM calls from
$O(|\mathcal{N}(v)|)$ per target to $O(1)$, enabling feasible end-to-end optimization.

\textbf{GNN message passing.}
Because target and neighbor nodes supply embeddings differently, message passing becomes inherently asymmetric.
A heterogeneous GraphSAGE layer computes:
\begin{equation}
\begin{aligned}
h_i^{(l+1)} &=
\sigma\!\Big(
W^{(l)} \cdot
\mathrm{AGG}_r\!\big(
\{h_i^{(l)}\} \cup \\[-3pt]
&\qquad\{h_j^{(l)} : v_j \in \mathcal{N}_r(v_i)\}
\big)
\Big)
\end{aligned}
\end{equation}
where $\mathrm{AGG}_r$ is relation-type mean aggregation, and $\sigma$ an activation function.  
The final embedding \(h_i^{(L)}\) integrates CoT semantics with relational dependencies.
The gradients are propagated only through the target embedding while neighbor embeddings serve as static messages. Since GNN aggregates cached neighbor embeddings, structural
information shapes target representations every iteration, while LLM parameters receive
gradients that are implicitly modulated by graph structure.  

\textbf{Training Objective.} 
We optimize the node-level binary cross-entropy loss to learn both the LoRA-adapted LLM encoder and the GNN:
\begin{equation}
\begin{aligned}
\mathcal{L} &= 
-\frac{1}{|V_{\text{train}}|} \sum_{v_i \in V_{\text{train}}} \left[ 
y_i \log f_\theta(v_i) \right. \\
&\quad \left. + (1 - y_i) \log (1 - f_\theta(v_i)) \right]
\end{aligned}
\end{equation}

To sum up, FraudCoT’s design aligns semantic reasoning and structural learning under a single optimization view.  
Distilled CoTs approximate the conditional explanation \(P(r_i | x_i, \mathcal{N}(v_i))\), while joint alignment preserves these explanations through message passing, leading to more calibrated predictions.  
Moreover, each node prediction is backed by a human-readable reasoning path, facilitating human diagnosis.

We further demonstrate the computational efficiency of our approach, and present a detailed complexity analysis in Appendix~\ref{sec:complexity}.

\section{Experiments}
We conduct extensive experiments to evaluate FraudCoT, including performance comparisons with strong baselines, component analyses, and sensitivity studies.

{
\setlength{\tabcolsep}{1mm}
\begin{table}[!t]
\centering
\resizebox{\linewidth}{!}{
\begin{tabular}{@{}ccccc@{}}
\toprule
Dataset & \# Nodes & \# Edges & \makecell{\# Edge \\ Types} & \# Train / Val / Test \\ \midrule
InstantVideo & 37,126 & 9,883,406 & 3 & 1,098 / 549 / 1,098 \\
DigitalMusic & 64,706 & 7,732,420 & 3 & 6,444 / 3,222 / 6,444 \\
PromotionAbuse & 371,464 & 1,388,598 & 3 & 143,592 / 17,949 / 17,949 \\ \bottomrule
\end{tabular}
}
\caption{Dataset statistics.}
\label{tab:dataset-stats}
\end{table}
}

{
\setlength{\tabcolsep}{1mm}
\begin{table*}[!t]
\centering
\resizebox{\textwidth}{!}{
\begin{tabular}{@{}c|cc|ccc|ccc|ccc@{}}
\toprule
\multirow{2}{*}{Method} & \multirow{2}{*}{LLM} & \multirow{2}{*}{GNN} & \multicolumn{3}{c|}{InstantVideo} & \multicolumn{3}{c|}{DigitalMusic} & \multicolumn{3}{c}{PromotionAbuse} \\
 &  &  & MacroF1 & AUROC & AUPRC & MacroF1 & AUROC & AUPRC & MacroF1 & AUROC & AUPRC \\ \midrule
MLP & \nomark & \nomark & \ms{76.84}{0.70} & \ms{84.56}{0.76} & \ms{71.48}{1.02} & \ms{81.13}{0.13} & \ms{90.44}{0.21} & \ms{76.44}{0.33} & \ms{57.49}{0.35} & \ms{71.08}{0.42} & \ms{58.12}{0.51} \\
RGCN & \nomark & \yesmark & \ms{78.78}{0.28} & \ms{86.67}{0.04} & \ms{74.79}{0.36} & \ms{81.67}{0.17} & \ms{91.33}{0.16} & \ms{79.49}{0.37} & \ms{66.78}{0.41} & \ms{79.49}{0.38} & \ms{67.70}{0.46} \\
HGT & \nomark & \yesmark & \ms{81.53}{0.62} & \ms{88.77}{0.31} & \ms{81.25}{1.13} & \ms{83.23}{0.16} & \ms{91.50}{0.08} & \ms{82.24}{0.29} & \ms{60.01}{0.47} & \ms{64.09}{0.54} & \ms{41.88}{0.62} \\
ConsisGAD & \nomark & \yesmark & \ms{81.69}{0.39} & {\ul \ms{89.12}{0.03}} & {\ul \ms{82.75}{0.27}} & {\ul \ms{83.27}{0.21}} & \ms{91.65}{0.08} & {\ul \ms{82.90}{0.32}} & {\ul \ms{75.56}{0.36}} & \ms{84.29}{0.41} & {\ul \ms{68.82}{0.48}} \\
PMP & \nomark & \yesmark & {\ul \ms{82.08}{0.29}} & \ms{88.97}{0.17} & \ms{80.16}{0.85} & \ms{80.41}{0.40} & \ms{82.51}{3.14} & \ms{64.32}{1.05} & \ms{58.89}{0.44} & \ms{62.93}{0.51} & \ms{40.45}{0.59} \\
GAAP & \nomark & \yesmark & \ms{79.06}{1.08} & \ms{86.90}{0.80} & \ms{74.90}{1.50} & \ms{82.30}{0.40} & {\ul \ms{91.78}{0.17}} & \ms{80.15}{0.50} & \ms{72.75}{0.38} & \ms{81.37}{0.46} & \ms{65.13}{0.53} \\ \midrule
LLM & \yesmark & \nomark & \ms{39.56}{0.42} & \ms{49.72}{0.39} & \ms{34.55}{0.48} & \ms{42.50}{0.61} & \ms{51.78}{0.57} & \ms{26.09}{0.69} & \ms{48.86}{0.71} & \ms{51.00}{0.80} & \ms{26.91}{0.93} \\
LLM-SFT & \yesmark & \nomark & \ms{65.75}{0.50} & \ms{70.09}{0.37} & \ms{51.51}{0.31} & \ms{63.79}{1.08} & \ms{70.13}{1.67} & \ms{45.80}{2.02} & \ms{49.12}{0.52} & \ms{55.39}{0.50} & \ms{35.75}{0.63} \\
InstructGLM & \yesmark & \nomark & \ms{65.29}{0.50} & \ms{69.59}{0.04} & \ms{52.09}{0.01} & \ms{70.39}{3.60} & \ms{78.52}{3.70} & \ms{54.30}{5.48} & \ms{54.61}{0.66} & \ms{54.09}{0.75} & \ms{34.75}{0.87} \\ \midrule
LLaGA & \yesmark & \yesmark & \ms{59.66}{1.79} & \ms{57.78}{2.30} & \ms{43.89}{2.26} & \ms{63.62}{0.79} & \ms{64.27}{1.27} & \ms{40.98}{1.02} & \ms{50.42}{1.11} & \ms{55.07}{1.36} & \ms{35.26}{1.48} \\
GraphGPT & \yesmark & \yesmark & \ms{60.20}{0.86} & \ms{63.16}{0.94} & \ms{49.87}{0.98} & \ms{60.27}{8.66} & \ms{62.23}{13.93} & \ms{41.80}{14.40} & \ms{52.68}{1.42} & \ms{58.92}{1.57} & \ms{38.47}{1.63} \\
HiGPT & \yesmark & \yesmark & \ms{60.91}{4.52} & \ms{66.06}{6.13} & \ms{48.78}{6.49} & \ms{60.20}{0.64} & \ms{65.00}{1.20} & \ms{39.42}{0.82} & \ms{55.33}{1.26} & \ms{60.25}{1.39} & \ms{40.21}{1.48} \\
TAPE & \yesmark & \yesmark & \ms{81.18}{0.58} & \ms{88.70}{0.21} & \ms{79.28}{0.58} & \ms{82.98}{0.17} & \ms{91.59}{0.05} & \ms{81.99}{0.19} & \ms{74.12}{0.33} & {\ul \ms{86.47}{0.42}} & \ms{67.21}{0.49} \\
FLAG & \yesmark & \yesmark & \ms{57.36}{0.18} & \ms{57.65}{0.18} & \ms{41.75}{0.05} & \ms{62.46}{0.34} & \ms{63.44}{0.26} & \ms{38.59}{0.23} & \ms{58.06}{0.37} & \ms{63.28}{0.41} & \ms{42.17}{0.52} \\ \midrule
FraudCoT & \yesmark & \yesmark & \textbf{$^*$\ms{83.21}{0.21}} & \textbf{$^*$\ms{90.73}{0.19}} & \textbf{$^*$\ms{84.10}{0.26}} & \textbf{$^*$\ms{84.23}{0.24}} & \textbf{$^*$\ms{93.08}{0.21}} & \textbf{$^*$\ms{84.49}{0.28}} & \textbf{$^*$\ms{77.95}{0.29}} & \textbf{$^*$\ms{88.03}{0.33}} & \textbf{$^*$\ms{73.65}{0.41}} \\ \bottomrule
\end{tabular}
}
\caption{Comparison of fraud detection performance (mean$\pm$std \%) across datasets. The best and second-best results are marked in bold and underlined, respectively. $^*$ indicates the improvement over the second-best method is significant with $p < 0.05$ via a paired t-test.}
\label{tab:results}
\end{table*}
}

\subsection{Experimental Setup}

\textbf{Datasets.}
We evaluate our method on three graph datasets, including two public benchmarks--InstantVideo and DigitalMusic--from the Amazon Reviews corpus~\cite{mcauley2013amateurs}, and a proprietary industrial dataset. InstantVideo and DigitalMusic are category-specific datasets, where each labeled node represents a user review associated with raw text and labeled as either helpful or unhelpful.
\bd{We further include one proprietary industrial dataset, referred to as PromotionAbuse, which is a real-world graph sampled from our industry partner, ByteDance. It contains real-world abusive behaviors targeting promotional incentives.}

Following prior work~\cite{dou2020enhancing}, we construct heterogeneous graphs with three types of edges for InstantVideo and DigitalMusic: reviews posted by the same user (R-U-R), reviews posted on the same product (R-P-R), and same-product reviews posted with the same rating and within the same week (R-S-R). 
Table~\ref{tab:dataset-stats} summarizes dataset statistics. Note that the total number of nodes in labeled splits can be smaller than the full graph, reflecting the real-world case where most nodes remain unlabeled.

\textbf{Baselines.}
We benchmark against a wide range of competitive models: 
(i) \textit{GNNs}, including GraphSAGE~\cite{hamilton2017inductive}, HGT~\cite{hu2020heterogeneous}, ConsisGAD~\cite{chen2024consistency}, PMP~\cite{zhuo2024partitioning}, and GAAP~\cite{duan2025global};
(ii) \textit{Graph-agnostic models}, including MLP~\cite{rosenblatt1958perceptron}, a Qwen3-8B LLM~\cite{qwen3technicalreport} and its finetuned version targeted for fraud detection; 
(iii) \textit{graph-enhanced LLMs}, including LLaGA~\cite{chen2024llaga}, GraphGPT~\cite{tang2024graphgpt}, HiGPT~\cite{tang2024higpt}, and InstructGLM~\cite{ye2023language};
and (iv) \textit{LLM-enhanced GNNs}, represented by TAPE~\cite{he2024harnessing} and FLAG~\cite{yang2025flag}.
All baselines are implemented using official code.

\textbf{Implementation Details.}
We conduct experiments using a Linux system with 64 Intel(R) Xeon(R) Gold 6346 CPUs, 1TB of RAM, and one NVIDIA A100 GPU (80GB).
For all LLM-tuning methods, we use the Qwen3-8B LLM backbone~\cite{qwen3technicalreport} for fair comparison. 
We apply LoRA~\cite{hu2022lora} to all attention layers and use AdamW~\cite{DBLP:conf/iclr/LoshchilovH19} optimizer for finetuning.
We adopt classification metrics including Macro-F1, AUROC, and AUPRC (detailed in Appendix~\ref{ap:metric}); and report the mean and standard deviation over 5 random seeds. 
The model is implemented via PyTorch~\cite{paszke2019pytorch} and DGL~\cite{wang2019deep}.

We detail the parameter settings in Appendix~\ref{sec:parameter}.

\subsection{Performance Evaluation}

Table~\ref{tab:results} reports the main results. Across all datasets and metrics, FraudCoT consistently outperforms GNN-only and LLM-enhanced baselines. 
\bd{The gains over the strongest competitors are especially significant on the industrial PromotionAbuse dataset, demonstrating real-world applicability.}
\begin{itemize}
\item Pure-LLM finetuning (LLM-SFT, InstructGLM) treats nodes independently and cannot propagate multi-hop evidence that characterizes collective fraud. Even with informative text, the lack of relational context yields brittle, overfit decisions and weaker recall on ambiguous cases.

\item Graph-enhanced LLMs enhance semantic analysis but often produce generic, template-like rationales. Without tight alignment between token cues and specific neighbors/edge types, they struggle to separate genuine community effects from spurious correlations in mixed homophily/heterophily settings.

\item LLM-enhanced GNNs such as TAPE gain from prompting yet remain confined by predefined templates, encouraging shallow cue extraction over deeper reasoning. 

\item FraudCoT unifies autonomous reasoning with structure-aware learning in a compute-efficient pipeline. Positive and negative CoT distillation suppresses misleading trajectories, and co-training aligns semantics with message passing, yielding stable gains across public and industrial datasets.
\end{itemize}

\subsection{Ablation Study}

\begin{figure}[!t]
  \centering
  \begin{subfigure}[b]{0.49\linewidth}
    \centering
    \includegraphics[width=\linewidth]{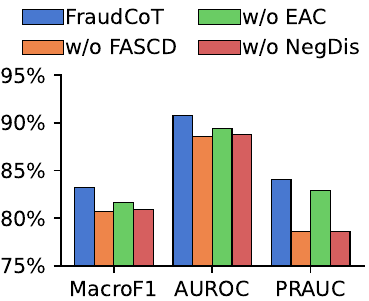}
    \caption{InstantVideo}
  \end{subfigure}
  \hfill
  \begin{subfigure}[b]{0.49\linewidth}
    \centering
    \includegraphics[width=\linewidth]{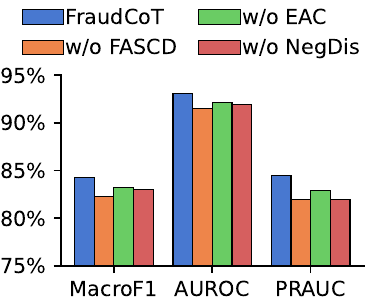}
    \caption{DigitalMusic}
  \end{subfigure}
  \caption{Ablation study on FraudCoT components.}
  \label{fig:ab}
\end{figure}

We further examine the contribution of each architectural component through three variants: (i) removing the Fraud-Aware Selective CoT Distillation stage (w/o FASCD), which eliminates distilled multi-hop rationales and reduces the model to relying solely on raw textual descriptions; (ii) disabling Efficient Asymmetric Co-training (w/o EAC), which freezes the LLM encoder and prevents semantic representations from adapting to relational feedback; and (iii) removing the suppression of incorrect reasoning trajectories (w/o NegDis), thereby allowing the student model to internalize spurious or weakly supported teacher rationales. As shown in Figure~\ref{fig:ab}, all variants yield consistent declines across datasets and metrics. w/o FASCD exhibits the most pronounced deterioration on ranking-oriented measures, indicating that distilled reasoning paths play a central role in capturing cross-neighborhood dependencies. w/o EAC leads to misaligned semantic and structural signals, degrading global calibration, while w/o NegDis reduces robustness by allowing implausible reasoning paths to influence the semantic space.

Overall, the results show that FraudCoT’s performance gains arise from the interaction between its reasoning, semantic, and structural components rather than from any single mechanism in isolation. Removing distilled CoTs weakens the model’s ability to form graph-grounded hypotheses, freezing the encoder disrupts the coupling between textual semantics and graph topology, and eliminating negative-path filtering increases sensitivity to misleading cues. The full framework jointly mitigates these failure modes, indicating that selective reasoning distillation and structure-aware co-training function as mutually reinforcing processes within a unified reasoning–graph learning paradigm.

\subsection{Effect of Distillation Samples}

\begin{figure}[!t]
  \centering
  \begin{subfigure}[b]{0.49\linewidth}
    \centering
    \includegraphics[width=\linewidth]{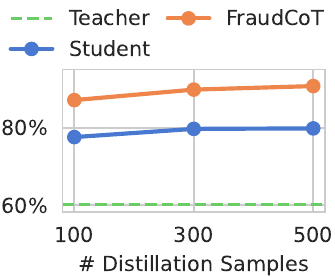}
    \caption{InstantVideo (AUROC)}
  \end{subfigure}
  \hfill
  \begin{subfigure}[b]{0.49\linewidth}
    \centering
    \includegraphics[width=\linewidth]{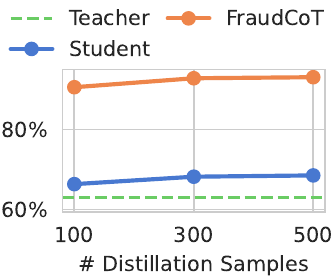}
    \caption{DigitalMusic (AUROC)}
  \end{subfigure}
\caption{Impact of the number of distillation samples on classification performance. \textit{Teacher} and \textit{Student} denote the performance achieved solely by Teacher and Student LLMs, respectively.}
  \label{fig:nsamples}
\end{figure}

To examine how much CoT supervision is needed, we vary the number of teacher-generated reasoning samples during Stage~1 distillation while keeping Stage~2 co-training fixed. As shown in Figure~\ref{fig:nsamples}, Stage~1 performance quickly saturates once around one hundred samples are provided. This suggests that the distilled student rapidly internalizes core reasoning patterns—detecting suspicious textual cues and linking them to neighbors—such that additional samples mainly introduce stylistic variation without noticeably improving standalone reasoning quality. We also observe that the zero-shot teacher LLMs frequently produce label-inconsistent rationales, whereas the distilled student LLMs yield more calibrated and faithful explanations.

In Stage~2, where the distilled student is jointly optimized with the GNN, larger distillation sets translate into steady accuracy gains. These results demonstrate that co-training acts as an amplifier of reasoning signals: even small improvements in CoT quality become reinforced through message passing and structural feedback. In practice, this means a modest number of annotated CoTs suffices to establish a strong baseline, while scaling distillation remains beneficial when paired with structure-aware learning.

\subsection{Effect of Negative Distillation Weight}

\begin{figure}[!t]
  \centering
  \begin{subfigure}[b]{0.49\linewidth}
    \centering
    \includegraphics[width=\linewidth]{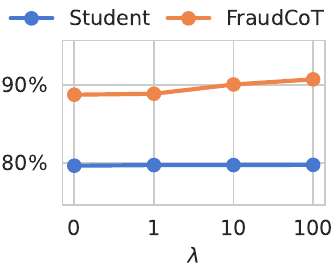}
    \caption{InstantVideo (AUROC)}
  \end{subfigure}
  \hfill
  \begin{subfigure}[b]{0.49\linewidth}
    \centering
    \includegraphics[width=\linewidth]{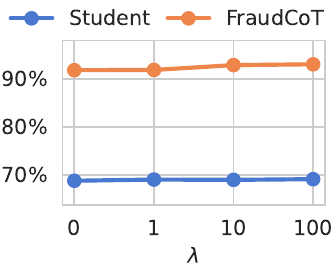}
    \caption{DigitalMusic (AUROC)}
  \end{subfigure}
\caption{Impact of $\mathcal{L}_{\text{UL}}$'s weight $\lambda$ on performance.}
  \label{fig:lambda}
\end{figure}

We study the effect of $\lambda$ in balancing positive imitation and negative suppression during reasoning transfer. Figure~\ref{fig:lambda} shows that increasing $\lambda$ consistently improves performance, especially after Stage~2 co-training where semantic–structural alignment magnifies reasoning quality. A crucial practical detail is that spurious rationales typically have low original token probabilities under the student model—meaning the unlikelihood loss $-\log(1-P_\theta)$ contributes only a weak gradient signal if $\lambda$ is small. Thus, a larger weight is required to make negative supervision competitive with positive imitation, effectively sharpening the model’s discrimination against seductive but incorrect reasoning paths.

The best overall performance is achieved at $\lambda = 100$, indicating that strong suppression of invalid trajectories is essential for fraud detection, where camouflage behaviors and noisy neighborhood cues are common. Without sufficiently penalizing incorrect explanations, the student tends to overfit teacher biases or rely on superficial correlations that do not persist across the graph. The results therefore highlight that negative-path distillation is not simply a regularizer but a core component for robust, graph-aware reasoning—supporting our argument that CoT distillation, when properly calibrated, is a more principled and resilient strategy than static prompt engineering in fraud scenarios.

\subsection{More Analysis}

Appendix \ref{sec:efficiency} provides efficiency comparison showing that our asymmetric co-training reduces training time while enabling much larger batch sizes than naive co-training. In addition, the case study in Appendix~\ref{sec:case} shows how FraudCoT performs semantic- and structure-aware CoT reasoning that differs from predefined prompting.

\section{Conclusion}

We introduce FraudCoT, a unified framework that distills chain-of-thought reasoning for graph-based fraud detection.
To enable autonomous graph reasoning beyond predefined prompting, we introduce fraud-aware selective CoT distillation that provides diverse reasoning paths for more reliable semantic–structural understanding.
To enable end-to-end LLM--GNN optimization for better semantic-structural alignment, we develop an efficient asymmetric co-training strategy that significantly reduces the computational cost of naive joint training.
Experiments on public and industrial datasets show improved analytic quality and accuracy, indicating that autonomous reasoning supervision outperforms predefined prompting for LLM-enhanced fraud detection.

\bibliography{custom}

\appendix

\section{Complexity Analysis}\label{sec:complexity}

\begin{table*}[!t]
\centering
\resizebox{\linewidth}{!}{
\begin{tabular}{@{}cccc@{}}
\toprule
\textbf{Method} & \textbf{Training Paradigm} & \textbf{Time Complexity} & \textbf{Space Complexity} \\ \midrule
Naive Co-train~\cite{liu2025scalable} & \makecell{End-to-end (Joint)} & \makecell[l]{$O(M N K^H L(TD^2 + T^2D)) + O(M N K^H DL)$} & $O(D^2L + N K^H (TD + D))$ \\
TAPE~\cite{he2024harnessing} & Two-stage & \makecell[l]{$O(M N L(TD^2 + T^2D)) + O(M N K^H DL)$} & $O(D^2L + N (TD + D))$ \\
STAR~\cite{liu2025scalable} & Two-stage & \makecell[l]{$O(M N L(TD^2 + T^2D)) + O(M N K^H DL)$} & $O(D^2L + N (TD + D))$ \\
FLAG~\cite{yang2025flag} & Alternating & \makecell[l]{$O(I N L(TD^2 + T^2D)) + O(I M N K^H DL)$} & $O(D^2L + N (TD + D))$ \\
FraudCoT (Ours) & End-to-end (Asymmetric) & \makecell{$O((US + N + MN) L(TD^2 + T^2D)) + O(M N K^H DL)$} & \makecell{$O(D^2L + N(TD + D))$} \\ \bottomrule
\end{tabular}
}
\caption{
Comparison of time and space complexity. 
($N$: number of nodes, $M$: number of epochs, $K$: neighbor sample size per hop, $H$: number of hops, $L$: number of LLM layers, $T$: token length, $D$: hidden dimension. 
$U$: number of distillation nodes, $S$: number of reasoning paths, $I$: number of alternating iterations.)
}
\label{tab:complexity}
\end{table*}

We analyze the computational complexities of the proposed method. Let $N$ be the number of nodes, $E$ the number of edges, $D$ the hidden dimension, $L$ the number of LLM or GNN layers, $K$ the number of sampled neighbors per hop, $H$ the number of sampled hops, $U$ the number of nodes used for distillation (each producing $S$ CoTs), and $T$ the token length of CoT-augmented text.

\textbf{Time Complexity.} A single LLM forward or backward pass on an input of length $T$ costs $O\!(L(TD^{2} + T^{2}D))$. In the distillation stage, teacher generation and student imitation require $US$ such LLM calls, giving a total cost of $O\!(USL(TD^{2} + T^{2}D))$. In the co-training stage, constructing the neighbor-embedding cache requires encoding all $N$ nodes once for $O\!(NL(TD^{2} + T^{2}D))$, after which every training epoch re-encodes each target node once for another $O\!(NL(TD^{2} + T^{2}D))$, while heterogeneous GraphSAGE message passing with $H$ hops and $K$ sampled neighbors per hop adds $O(NK^{H}DL)$ per epoch. For $M$ epochs, co-training thus costs $O\!(N \cdot L(TD^{2} + T^{2}D) + M(N \cdot L(TD^{2} + T^{2}D) + NK^{H}DL))$. Therefore, the overall complexity of FraudCoT is $O\!((US + N + MN)\,L(TD^{2} + T^{2}D) + MNK^{H}DL)$. Under fixed $T, D, L, K, H, S$ with bounded $U/N$, the total runtime grows linearly with $N$.

\textbf{Space Complexity.} The space cost of FraudCoT comes from LLM parameters, node-level representations, and the neighbor-embedding cache. The LLM contributes $O(D^{2}L)$ parameters. For node storage, CoT-augmented texts require $O(NT)$, while caching LLM hidden states costs $O(NTD)$; GNN representations add another $O(ND)$ per layer. The cache stores a node embedding for each node, contributing $O(ND)$. Distillation additionally stores $US$ CoTs and their hidden states, i.e., $O(US(T + TD))$, but this is typically smaller than full-graph caching. Hence, the peak memory usage is dominated by LLM hidden-state caching and neighbor embeddings, yielding a total space complexity of $O\!\big(D^{2}L + N(TD + D)\big)$, which grows linearly with $N$ under fixed $T, D, L, K, H$.

As summarized in Table~\ref{tab:complexity}, compared with naive end-to-end co-training that encodes both target and all $H$-hop sampled neighbors with the LLM in every epoch---incurring $O\!(MNK^{H}L(TD^{2} + T^{2}D))$ time and $O\!(NK^{H}TD)$ space for neighbor hidden states---our asymmetric design reduces LLM calls from $O(NK^{H})$ to $O(N)$ per epoch and replaces neighbor-side computation with lightweight GNN aggregation $O(MNK^{H}DL)$ and a compact cache $O(ND)$, thereby eliminating the dominant exponential LLM cost while preserving the same neighborhood coverage. 
In addition, FraudCoT achieves efficiency comparable to decoupled baselines (e.g., TAPE, FLAG, and STAR), while uniquely maintaining the advantage of end-to-end semantic--structural alignment.

\section{Evaluation Metrics}\label{ap:metric}

We evaluate model performance using standard classification metrics implemented in the scikit-learn library~\cite{scikit-learn}.

\textbf{Macro-F1}
To reduce the impact of class imbalance, we use the macro-averaged F1 score. For each class $c \in \{0,1\}$, let $\mathrm{TP}_c$, $\mathrm{FP}_c$, and $\mathrm{FN}_c$ denote the true positives, false positives, and false negatives when treating class $c$ as the positive class. The precision, recall, and F1 score for class $c$ are defined as $\mathrm{Precision}_c = \mathrm{TP}_c / (\mathrm{TP}_c + \mathrm{FP}_c)$, $\mathrm{Recall}_c = \mathrm{TP}_c / (\mathrm{TP}_c + \mathrm{FN}_c)$, and $\mathrm{F1}_c = 2 \cdot \mathrm{Precision}_c \cdot \mathrm{Recall}_c / (\mathrm{Precision}_c + \mathrm{Recall}_c)$, respectively. The Macro-F1 score is computed as $\mathrm{Macro\text{-}F1} = (\mathrm{F1}_0 + \mathrm{F1}_1)/2$.

\textbf{Area Under the ROC Curve (AUROC)}  
The ROC curve characterizes the relationship between the true positive rate (TPR) and the false positive rate (FPR) across varying decision thresholds. Given a threshold $t$, the true positive rate is $\mathrm{TPR}(t) = \mathrm{TP}(t) / (\mathrm{TP}(t) + \mathrm{FN}(t))$ and the false positive rate is $\mathrm{FPR}(t) = \mathrm{FP}(t) / (\mathrm{FP}(t) + \mathrm{TN}(t))$. The AUROC corresponds to the area under this curve and is expressed as $\mathrm{AUROC} = \int_0^1 \mathrm{TPR}(\mathrm{FPR}^{-1}(x)) \, dx$.

\textbf{Area Under the Precision–Recall Curve (AUPRC)}  
The precision–recall curve plots precision as a function of recall across classification thresholds. AUPRC is defined as $\mathrm{AUPRC} = \int_0^1 \mathrm{Precision}(r) \, dr$, where $r$ denotes recall. In highly imbalanced scenarios, AUPRC offers a more informative assessment than AUROC by concentrating on the performance for the positive class.

\section{Parameter Settings}\label{sec:parameter}
Following prior work~\cite{chen2024consistency}, we tune all hyperparameters using grid search based on validation performance in AUROC for the evaluated models. 
For the proposed method, we set the number of teacher CoTs $S=5$ from $U\in\{100,300,500\}$ nodes, and the unlikelihood weight $\lambda\in\{0.1,10,100\}$, the sampled neighbor size per hop $K=10$, hidden dimension $D=128$, and learning rate $\mu\in\{1\mathrm{e}{-5},1\mathrm{e}{-4},1\mathrm{e}{-3}\}$.
For LoRA-based finetuning of LLM methods, we tune the LoRA rank $r \in \{4, 8, 16, 32\}$, the LoRA dropout rate $\in \{0.0, 0.05, 0.1\}$, and the learning rate $\in \{1\text{e}{-5}, 3\text{e}{-5}, 1\text{e}{-4}\}$. 
We set the batch size to $128$ and finetune for up to $300$ epochs with early stopping based on validation loss. 
For all baseline methods, we tune hyperparameters within the recommended ranges reported in their original papers to ensure fair and optimized comparisons.

\section{Efficiency Evaluation}\label{sec:efficiency}

Table~\ref{tab:efficiency} compares naive and asymmetric co-training in terms of per-epoch time and the largest batch size that can fit on a single GPU. Across all three datasets, the proposed efficient asymmetric co-training scheme consistently reduces the training time by up to three orders of magnitude compared with naive co-training, especially on the largest industrial graph, where naive joint optimization becomes practically infeasible. 

Meanwhile, the maximal batch size increases from only a few instances per step to hundreds, indicating substantially better memory utilization and allowing stable optimization with standard batch sizes. These comparisons confirm that the proposed asymmetric co-training design turns end-to-end LLM--GNN training from a theoretically desirable but computationally prohibitive objective into a practically viable procedure, while still retaining the performance gains over decoupled LLM-enhanced GNN baselines shown in Table~\ref{tab:results}.

\begin{table}[t]
\centering
\resizebox{\linewidth}{!}{
\begin{tabular}{@{}cccc@{}}
\toprule
Dataset & \makecell{Naive\\Co-train} & \makecell{Asymmetric\\Co-train (Ours)} & \\ \midrule
&\multicolumn{2}{c}{Time per Epoch (s)} & \multicolumn{1}{c}{Speedup} \\ \midrule
InstantVideo     & 7{,}352    & 9    & 817$\times$ \\
DigitalMusic     & 58{,}639   & 55   & 1{,}066$\times$ \\
PromotionAbuse   & 251{,}206  & 261  & 962$\times$ \\ \midrule
&\multicolumn{2}{c}{Maximum Batch Size} & \multicolumn{1}{c}{Increase} \\ \midrule
InstantVideo     & 4     & 256  & 64$\times$ \\
DigitalMusic     & 2     & 256  & 128$\times$ \\
PromotionAbuse   & 1     & 128  & 128$\times$ \\
\bottomrule
\end{tabular}
}
\caption{Efficiency comparison between naive and asymmetric co-training, reported in time per epoch and maximum batch size.}
\label{tab:efficiency}
\end{table}

\section{Case Study}\label{sec:case}

\begin{figure*}[tp]
\centering
\begin{tcolorbox}[
    colback=gray!5,
    colframe=black!20,
    fonttitle=\bfseries\color{black},
    title={Prompt Template of FraudCoT},
    width=\linewidth
]
\textbf{<System Message>}

You are provided with a list of Amazon customers’ reviews. Each review is
classified as either helpful or unhelpful based on their interactions and content.

\textbf{<User Message>}

Target Node: [TARGET\_TEXT]

Neighbors: [NEIGHBOR\_TEXTS]

1. Give 1–2 short reasoning points for the unhelpfulness of the target node itself (each within 20 words). If none, just say ``Looks normal''.

2. Give 1–2 short reasoning points for connections inferring unhelpfulness (each within 20 words). If none, just say ``Looks normal''.

3. Give a prediction (Helpful / Unhelpful).
\end{tcolorbox}
\caption{Prompt template of FraudCoT for analytical text generation.}
\label{fig:fraudcot-template}
\end{figure*}

\begin{figure*}[tp]
\centering
\begin{tcolorbox}[
    colback=gray!5,
    colframe=black!20,
    fonttitle=\bfseries\color{black},
    title={Prompt Template of FLAG}
]
\textbf{<System Message>}

You are provided with a list of Amazon customers’ reviews. Each review is classified as either helpful or unhelpful based on their interactions and content.

\textbf{<User Message>}

Target Node: [TARGET\_TEXT]

Neighbors: [NEIGHBOR\_TEXTS]

1. Your task is to generate a brief discriminative text for the target review that directly relates to distinguishing it as either helpful or unhelpful. Ensure that the generated text highlights specific features that help differentiate the review’s classification while avoiding any general or irrelevant information.

2. Your task is to generate a brief residual text for the target review that captures the portions of its text that do not directly relate to distinguishing it as either helpful or unhelpful. The residual text should highlight non-discriminative features that are not helpful in differentiating the review’s classification. Focus on generating text that includes general background details, unrelated observations, or information that does not influence the classification task.
\end{tcolorbox}
\caption{Prompt template of FLAG for analysis generation.}
\label{fig:flag-template}
\end{figure*}

\begin{figure*}[tp]
\centering
\begin{tcolorbox}[
    colback=gray!5,
    colframe=black!20,
    fonttitle=\bfseries\color{black},
    title={Ego-Graph Description}
]
\textbf{<Target Node>}

Rating: 5.0; Text: Amazon Unbox rocks!! The Unbox service is easy to use, easy to download, and you get your episodes a day after they air! Mac messed up again and thought their service was going to be the only one on the block...

\textbf{<Neighbor Nodes>}

Relation: Same Rating; Rating: 5.0; Text: Another great season...

Relation: Same Product; Rating: 4.0; Text: OVERALL Good season...

Relation: Same User; Rating: 1.0; Text: Almost all shows on FOX are garbage...

Relation: Same User; Rating: 1.0; Text: I actually stopped watching this show...
\end{tcolorbox}
\caption{Ego-graph data in case study.}
\label{fig:subgraph}
\end{figure*}

\begin{figure*}[tp]
\centering
\begin{tcolorbox}[
    colback=gray!5,
    colframe=black!20,
    fonttitle=\bfseries\color{black},
    title={Analysis Generated by FraudCoT}
]
1. Target node unhelpfulness: Very short, promotional-sounding praise without specific details about the product’s performance...

2. Connection unhelpfulness: Contrast between the overly glowing 5-star review and multiple hostile 1-star reviews hints at unreliable reviewing behavior...

3. Prediction: Unhelpful
\end{tcolorbox}
\caption{Analysis generated by FraudCoT in case study.}
\label{fig:fraudcot-response}
\end{figure*}

\begin{figure*}[tp]
\centering
\begin{tcolorbox}[
    colback=gray!5,
    colframe=black!20,
    fonttitle=\bfseries\color{black},
    title={Analysis Generated by FLAG}
]
1. Discriminative Text

The reviewer clearly describes specific service features—easy downloading, simple use, and next-day episode availability—providing concrete reasons for their positive rating. These details directly inform potential buyers about the reliability and convenience of Amazon Unbox...

2. Residual Text

General enthusiasm about being “very happy” with the service and the aside about “Mac messed up again” add personal sentiment and commentary unrelated to evaluating the usefulness of the review for buyers...
\end{tcolorbox}
\caption{Analysis generated by FLAG in case study.}
\label{fig:flag-response}
\end{figure*}

To reveal how FraudCoT differs from predefined prompting, we present a case study on the InstantVideo dataset. The prompt templates of FraudCoT and FLAG~\cite{yang2025flag}, a representative baseline method, are provided in Figures~\ref{fig:fraudcot-template} and \ref{fig:flag-template}, respectively. We randomly select an unhelpful review and its ego-graph, which is described in Figure~\ref{fig:subgraph}. We compare the analyses produced by FraudCoT and FLAG, respectively. 

Following the prompt templates, FraudCoT produces a two-stage reasoning path that (i) diagnoses suspicious properties of the target review and (ii) ties the decision to connection-level signals across the ego-graph. FLAG, in contrast, outputs discriminative and residual segments of the target review. We make the following observations on their generated analyses in Figures~\ref{fig:fraudcot-response} and \ref{fig:flag-response}, respectively:
\begin{itemize}
    \item FraudCoT explicitly leverages cross-review behavioral anomalies. In comparison, FLAG receives the complete ego-graph but focuses mainly on discriminative phrases from the target review itself, failing to incorporate relational signals such as user-level inconsistencies or cross-review patterns.
    \item Through positive and negative CoT distillation, FraudCoT learns to suppress label-inconsistent rationales and produce label-faithful reasoning. In contrast, FLAG’s discriminative and residual texts remain fluent but tend to stay close to the target review’s original wording, offering less emphasis on underlying patterns.
\end{itemize}

\end{document}